\documentclass{esannV2}
\usepackage[dvips]{graphicx}
\usepackage[latin1]{inputenc}
\usepackage{amssymb,amsmath,array}
\usepackage{hyperref}
\usepackage[linesnumbered, ruled, vlined]{algorithm2e}
\usepackage{caption}
\usepackage{subcaption}
\usepackage{tikz-cd}
\usetikzlibrary{arrows, matrix}
\tikzcdset{every label/.append style = {scale=1},
every matrix/.append style={nodes={font=\LARGE}}}
\usepackage{adjustbox}
\newcommand{\bs}{\boldsymbol}

\newcommand{\mcl}{\mathcal}
\newcommand{\mrm}{\mathrm}

\begin{document}
\title{A Bayesian Variational Principle for Dynamic Self Organizing Maps}

\author{Anthony Fillion$^1$ and Thibaut Kulak$^1$ and François Blayo$^{1}$
%
%
\vspace{.3cm}\\
%
1- NeoInstinct - NeoLab \\
3 rue Traversière, Lausanne - Switzerland
%
}

\maketitle

\begin{abstract}
  We propose organisation conditions that yield a method for training SOM with
  adaptative neighborhood radius in a variational Bayesian framework. This
  method is validated on a non-stationary setting and compared in an
  high-dimensional setting with an other adaptative method.
\end{abstract}

\section{Introduction}
Self Organizing Maps (SOM, \cite{kohonen2012self}) is a biologically inspired
unsupervized vector quantization method for data modelization and visualization.
A map is made of a collection of weights $\left\{w_{z}\right\}_{z}$ indexed
by a discrete collection of points $z \in \left\{z_1, z_2, \ldots, z_n\right\}$.
The weights (or neurons) are meant to modelize some observation random variable
$\bs x$. The points are usually regularly placed on a two dimensional grid. In
this context, the SOM objective is to find representative weights with a well
organized response to stimuli i.e. weights specialize their response to some
kind of stimuli and weigths close in the grid space are close in the observation
space.

The original way to train such SOM is with Kohonen's iterations. Despite its
simplicity, they require a time dependant neighborhood function whose
neighborhood radius (or temperature) strictly decreases towards zero during
training. In \cite{berglund2006} and \cite{rougier2011}, heuristics are proposed
in a non-probabilistic framework to adapt this temperature. They lead to
``dynamic'' SOM algorithms that are able to track non-stationary datasets. In
\cite{verbeek2005}, the variational EM framework is used to design a cost
function for the SOM but the temperature parameter is still decreased as in
Kohonen. In \cite{bishop1998}, the topologically preserving properties of the
map stems from the basis function regularity. Hence those basis functions have
to be carefully chosen and their number must typically grow exponentially with
the observation space dimension.

We propose quantitative organization conditions that lead to an adaptative
temperature choice in a variational Bayesian framework. High dimensional
experiments show that our method is less sensitive to elasticity and outperforms
that of \cite{rougier2011}.

\section{Evidence Lower BOund}
In \cite{verbeek2005}, the neighborhood function is interpreted as the
variational posterior in a Expectation-Minization (EM, \cite{bishop2006pattern})
procedure. Therefore, they propose to minimize the following (opposite) ELBO to
train SOMs
\begin{align}\label{eq:F}
  F\left(\rho, \theta\right) &= \int \int \ln\left(\frac{q_{\bs z|\bs x}^{\rho}}{p_{\bs x, \bs z}^{\theta}}\right)q_{\bs z|\bs x}^{\rho}\mrm d \bs z p_{\bs x} \mrm d \bs x \\
                             &=  \underbrace{\int \mcl D\left(q_{\bs z|\bs x}^{\rho}, p_{\bs z|\bs x}^{\theta}\right)p_{\bs x}\mrm d\bs x}_{\text{organization}} + \underbrace{\mcl D\left(p_{\bs x}, p_{\bs x}^{\theta}\right)}_{\text{modelization}},
\end{align}
where, the data true density is $p_{\bs x}$ and the discrete latent variable is
$\bs z$. The complete data model
$p_{\bs x, \bs z}^{\theta} \propto e^{-\frac{1}{2}\left|\frac{\bs x - w_{\bs z}}{\sigma}\right|^{2}}$
is chosen Gaussian and the variational (amortized, homoskedastic) posterior
$q_{\bs z|\bs x}^{\rho} \propto e^{-\frac{1}{2}\left| \frac{\bs z - \mu_{\bs x}}{\lambda}\right|^{2}}$
as well. Their parameters are respectively
$\theta = \left\{z \mapsto w_{z}, \sigma\right\}, \rho = \left\{x \mapsto \mu_{x}, \lambda \right\}$.
The decomposition in the second equation reveals a modelization term that is the
Kullback-Leibler (KL) divergence between the data
density and the model marginal. As well as an other term that is the expected KL
divergence between the variational posterior and the model posterior (or
response to stimulus). The core idea is to choose the variational posterior
organized so that this term will favor organized posterior models.

\section{Organization conditions}
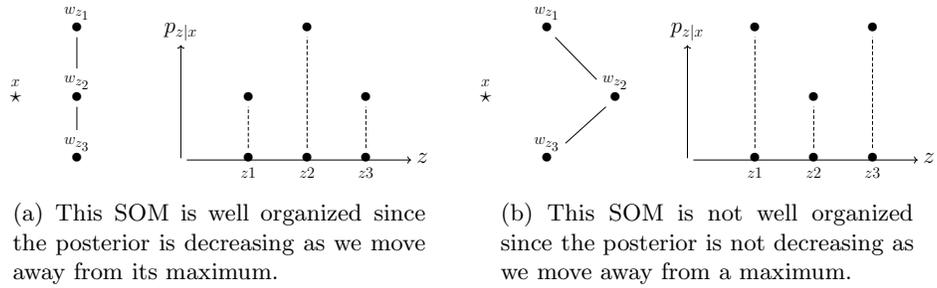
\begin{figure}[p]
\centering
\begin{subfigure}{.45\textwidth}
  \centering
  \adjustbox{scale=.5,center}{%
  \begin{tikzcd}
    & {\overset{w_{z_1}}{\bullet}} && {p_{z|x}} && \bullet \\
    {\overset{x}{\star}} & {\overset{w_{z_2}}{\bullet}} &&& \bullet && \bullet \\
    & {\overset{w_{z_3}}{\bullet}} && {} & {\underset{z1}{\bullet}} & {\underset{z2}{\bullet}} & {\underset{z3}{\bullet}} & z
    \arrow[no head, from=1-2, to=2-2]
    \arrow[no head, from=3-2, to=2-2]
    \arrow[from=3-4, to=1-4]
    \arrow[from=3-4, to=3-8]
    \arrow[dashed, no head, from=3-5, to=2-5]
    \arrow[dashed, no head, from=3-6, to=1-6]
    \arrow[dashed, no head, from=3-7, to=2-7]
  \end{tikzcd}}
\caption{This SOM is well organized since the posterior is decreasing as we
  move away from its maximum.}
\end{subfigure}\hspace{1cm}%
\begin{subfigure}{.45\textwidth}
  \centering
  \adjustbox{scale=.5,center}{%
  \begin{tikzcd}
    & {\overset{w_{z_1}}{\bullet}} && { p_{z|x}} & \bullet && \bullet \\
    {\overset{x}{\star}} && {\overset{w_{z_2}}{\bullet}} &&& \bullet \\
    & {\overset{w_{z_3}}{\bullet}} && { } & {\underset{z1}{\bullet}} & {\underset{z2}{\bullet}} & {\underset{z3}{\bullet}} & {z }
    \arrow[no head, from=1-2, to=2-3]
    \arrow[no head, from=3-2, to=2-3]
    \arrow[from=3-4, to=1-4]
    \arrow[from=3-4, to=3-8]
    \arrow[dashed, no head, from=3-5, to=1-5]
    \arrow[dashed, no head, from=3-6, to=2-6]
    \arrow[dashed, no head, from=3-7, to=1-7]
  \end{tikzcd}}
\caption{This SOM is not well organized since the posterior is not decreasing
  as we move away from a maximum.}
\end{subfigure}
\caption{On the left part of each subfigure, an observation $x$ and a SOM with
  1D neighborhood vizualised in its 2D observation space. On the right part, the
  graph of its posterior.}
\label{fig:org}
\end{figure}

Given an observation $\bs x$, let $w_{z^{*}_{\bs x}}$ be its closest weight. If
the distance to this weight $\left|w_{z^{*}_{\bs x}} - w_{\bs z}\right|$ in
observation space is increasing with the distance
$\left|z^{*}_{\bs x} - \bs z \right|$ in point space then the map is organized.
This mean that if the Bayesian posterior (or response)
$p_{\bs z | \bs x}^{\theta} \propto e^{-\frac{1}{2}\left|\frac{\bs x - w_{\bs z}}{\sigma}\right|^{2}}$
is plotted as a function of $\bs z$, it would be strictly decreasing around its
maximum. This is illustrated in Fig. \ref{fig:org}. Hence we
propose the following organization conditions:
\begin{description}
  \item[1:] The response $p_{\bs z|\bs x}^\theta$ has a dominating
        mode at $z_{\bs x}^{*} = \arg\min_{z} \left| \bs x - w_{z}\right| $.
  \item[2:] $\left|w_{\bs z} - w_{z^{*}_{\bs x}}\right| = \frac{1}{\eta} \left| \bs z - z^*_{\bs x}
  \right|$ when $\bs z$ is in the neighborhood of $z^*_{\bs x}$.
\end{description}
The first condition ensures that each weight specializes on some kind of
stimulus. The second one ensures that the weight distance is proportional to the
point distance by a scale $\eta$. In other words, the map preserves distances
locally.

\subsection{Variational posterior selection}
We now choose $\rho$ such that the variational posterior
$q_{\bs z|\bs x}^{\rho}$ verifies the organization conditions. Because the
variational posterior has a unique mode at $\mu_{\bs x}$,
choosing $$\mu_{\bs x} = z^*_{\bs x}$$ (like Kohonen does) favors reponses that
verifiy the first order organization condition. We also have,
\begin{align*}
  -\ln p_{\bs z|\bs x}^{\theta} = \frac{1}{\sigma^{2}}\langle \bs x - w_{z^{*}_{\bs x}} | w_{z^{*}_{\bs x}} - w_{\bs z}\rangle + \frac{1}{2}\left|\frac{w_{\bs z} - w_{\bs z^{*}_{\bs x}}}{\sigma}\right|^{2} + c,
\end{align*}
whith $c$ being a constant w.r.t $\bs z$. Because $z^{*}_{\bs x}$ is a mode, the
scalar product is almost null for $\bs z$ in the neighborhood of
$z^{*}_{\bs x}$. Therefore, if $\bs z$ is in the neighborhood of
$z_{\bs x}^{*}$, the organization term will favor responses such that
$\left|\frac{w_{z} - w_{z^{*}}}{\sigma}\right| \simeq \left|\frac{z - z^*}{\lambda} \right|$
. Thus choosing $$\lambda = \eta\sigma,$$ favors responses satisfying the second
organization condition.

\subsection{Stochastic Gradient calculations}
With the previous selection for $\rho$
($\mu_{\bs x} = z^{*}_{\bs x}, \lambda = \eta \sigma$), the objective $F$ in
Eq.(\ref{eq:F}) only depends on $\theta$. Algorithm \ref{alg:alg} computes a
stochastic approximation over $\bs x$ of its derivatives along
$\theta = \left\{w, \sigma\right\}$ assuming that
$z^{*}_{\bs x} = \arg\min_{z}\left|\bs x - w_{\bs z}\right|$ has no gradient.
Then any stochastic gradient descent method can be used to minimize $F$. A Python implementation is available at: \url{github.com/anthony-Neo/VDSOM}.

\begin{algorithm}
\KwData{$\eta$: elasticity\;
    $Z = \left\{z_i\right\}_{1\leq i \leq n}$: grid points\;
    $\left\{w_{z}\right\}_{z\in Z}, \sigma$: weights and variance\;
    $x, m$: observation and observation space dimension}
\KwResult{$g_{\sigma},\, \left\{g_{w_{z}}\right\}_{z\in Z}$ the stochastic gradients of $F$}
\SetKwFunction{softmax}{softmax}
\For{$z, y \in Z\times Z$}{
  $d_{z, y} := \left| z - y \right|^2$\;
  $f_z := \left|x - w_z\right|^2$\;
  $ \ln p_z := -m\ln\sigma - \frac{f_z}{2\sigma^2}$\;}
$z^* := \arg\min_{z} f_{z}$\;
$q_{z} := \softmax{$-\frac{d_{z, z^{*}}}{2\eta^2\sigma^2}$}$\;
$d^* := \sum_z d_{z, z^*}\times q_z$\;
$g_\sigma := \frac{m}{\sigma} + \frac{1}{\sigma^3}\sum_z \left[\eta\left(1 + \ln q_z - \ln p_z\right)\left(d_{z, z^*} - d^*\right) - f_z \right]q_z$\;
\For{$z\in Z$}{
    $g_{w_z} := - \frac{q_z}{\sigma^2}\left(x - w_z\right)$\;}
\caption{Stochastic gradient of $F$}\label{alg:alg}
\end{algorithm}

\section{Numerical Experiments}
\subsection{Non-stationary distributions}
\begin{figure}[p]
\centering
\includegraphics[width=\textwidth]{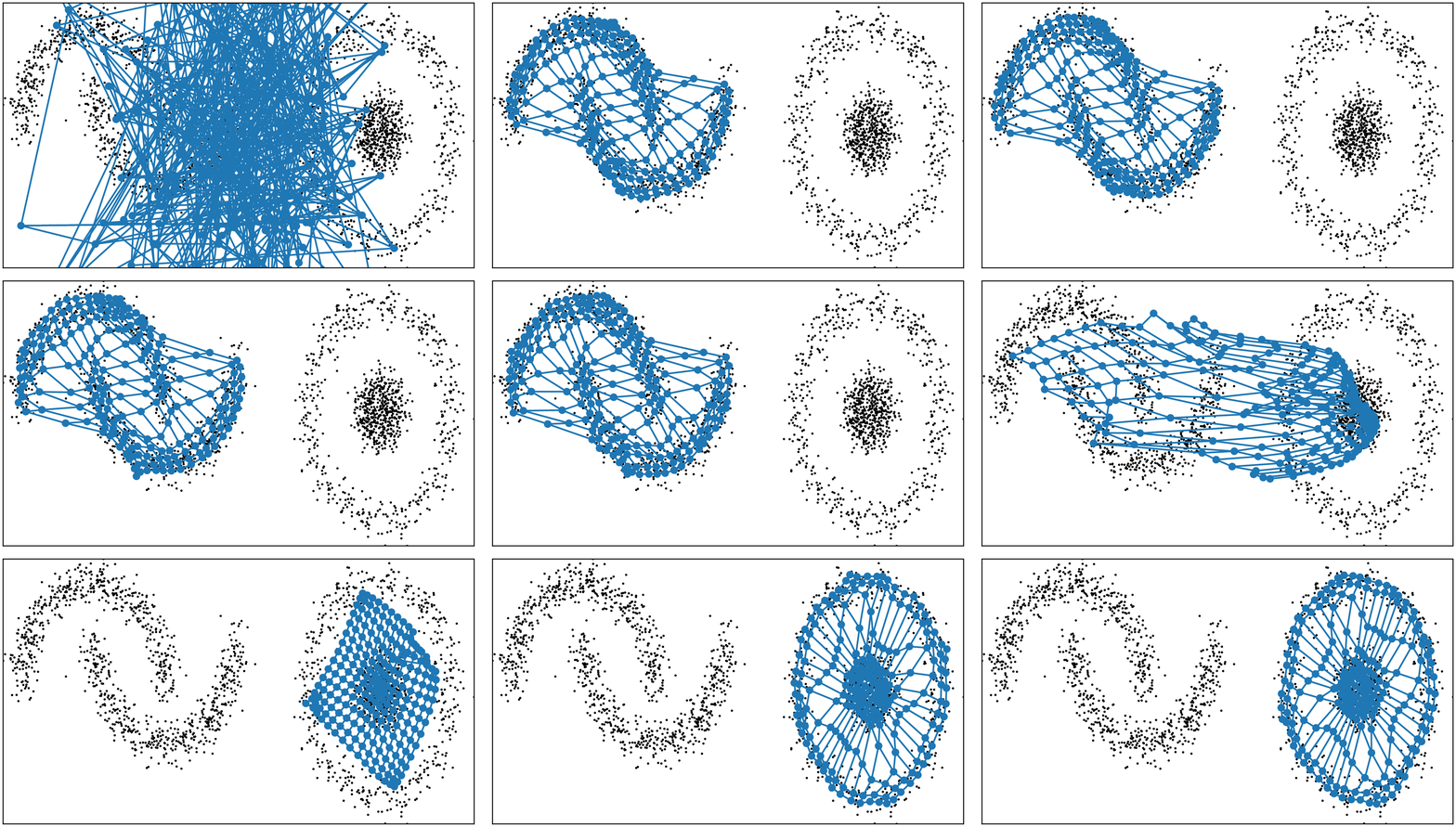}
\caption{Visualization of the map during the iterations, on a changing dataset}
\label{mutate}
\centering
\includegraphics[width=.8\textwidth]{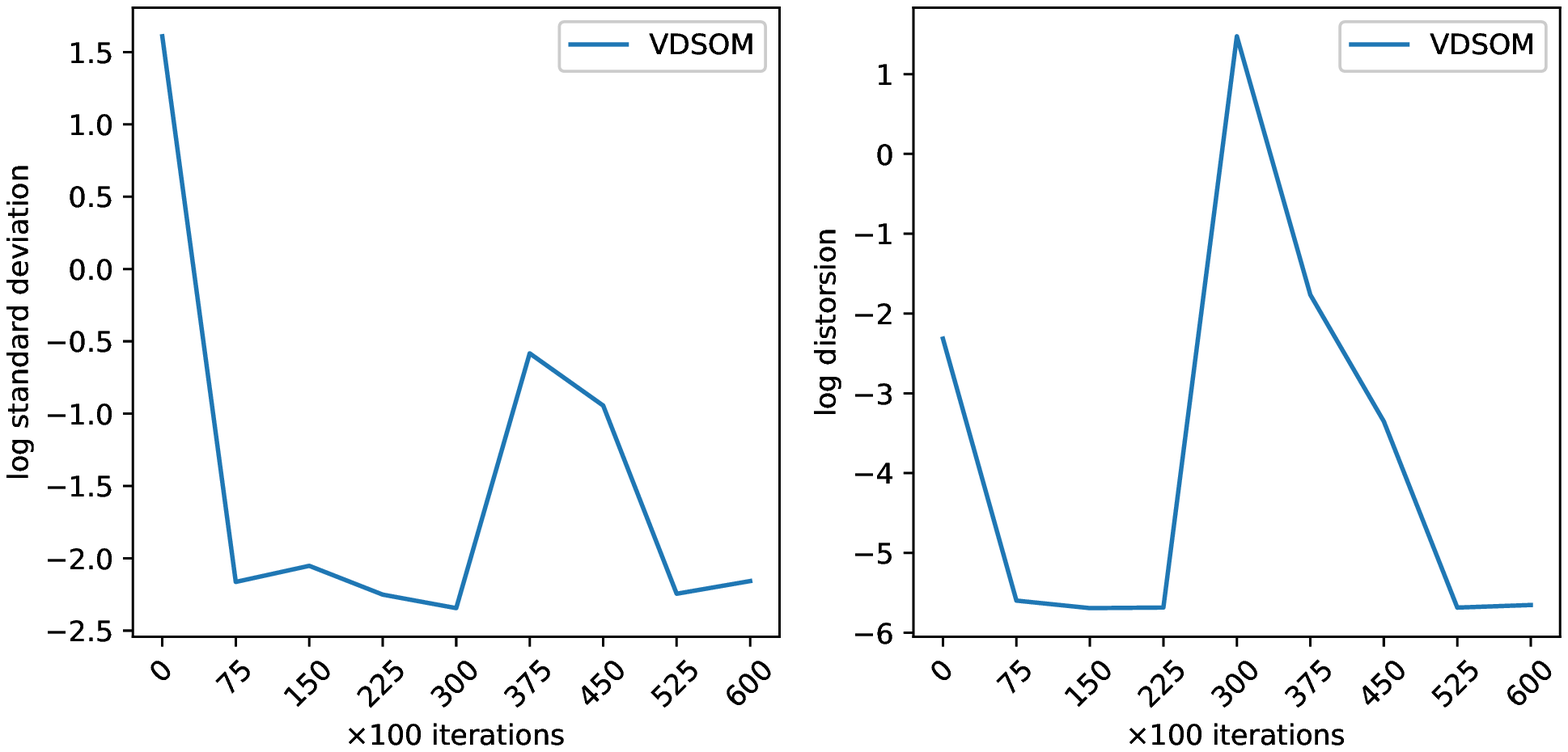}
\caption{Graphs of $\sigma$ and distorsion during the iterations, on a changing
  dataset}
\label{graphs}
\end{figure}

We used $8\times 7500$ steps of the Adam stochastic optimizer \cite{kingma2014}
with a learning rate of $\alpha = 10^{-3}$. The variance parameter is
initialized at $\sigma_{0} = 5$ and the initial weights sample a normal
Gaussian. The grid is a $15\times 15$ node lattice with regularly spaced points
between $-1, 1$ and the elasticity parameter is $\eta=1$. During the first half
of the iterations, the ``moons'' data set from sci-kit \cite{scikit} learn is
sampled. Then, during the second half, the ``circles'' data set is sampled. Step
$0$ and each $7500$ steps, data, weights and edges are plotted in
Fig.~\ref{mutate} from the upper left corner to the lower right one. We see that
the SOM tracks the changing data set, it correctly fits the observation density
(rather than its support as in \cite{rougier2011}) and it preserves the grid
neighborhood. In Fig.~\ref{graphs}, the log standard deviation $\sigma$ and the
log distorsion (mean over the samples of the min squared distance with the
weights, cf \cite{rougier2011}) are plotted against time. Spikes appear half
time which means that the method detects the change in data and adapts
neighborhood radius accordingly.

\subsection{High-dimensional distributions}
\begin{figure}[p]
\centering
\includegraphics[width=.8\textwidth]{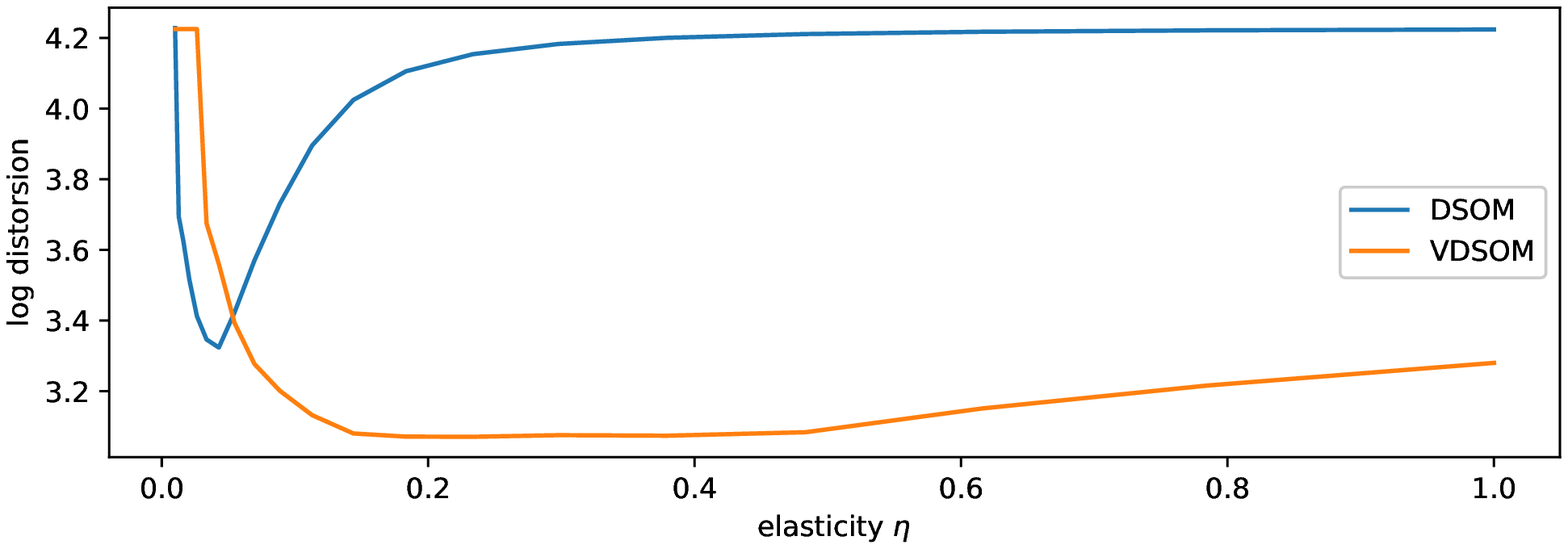}
\caption{sensitivity}
\label{fig:sensitivity}
\centering
\begin{subfigure}{.5\textwidth}
  \centering
  \includegraphics[width=\linewidth]{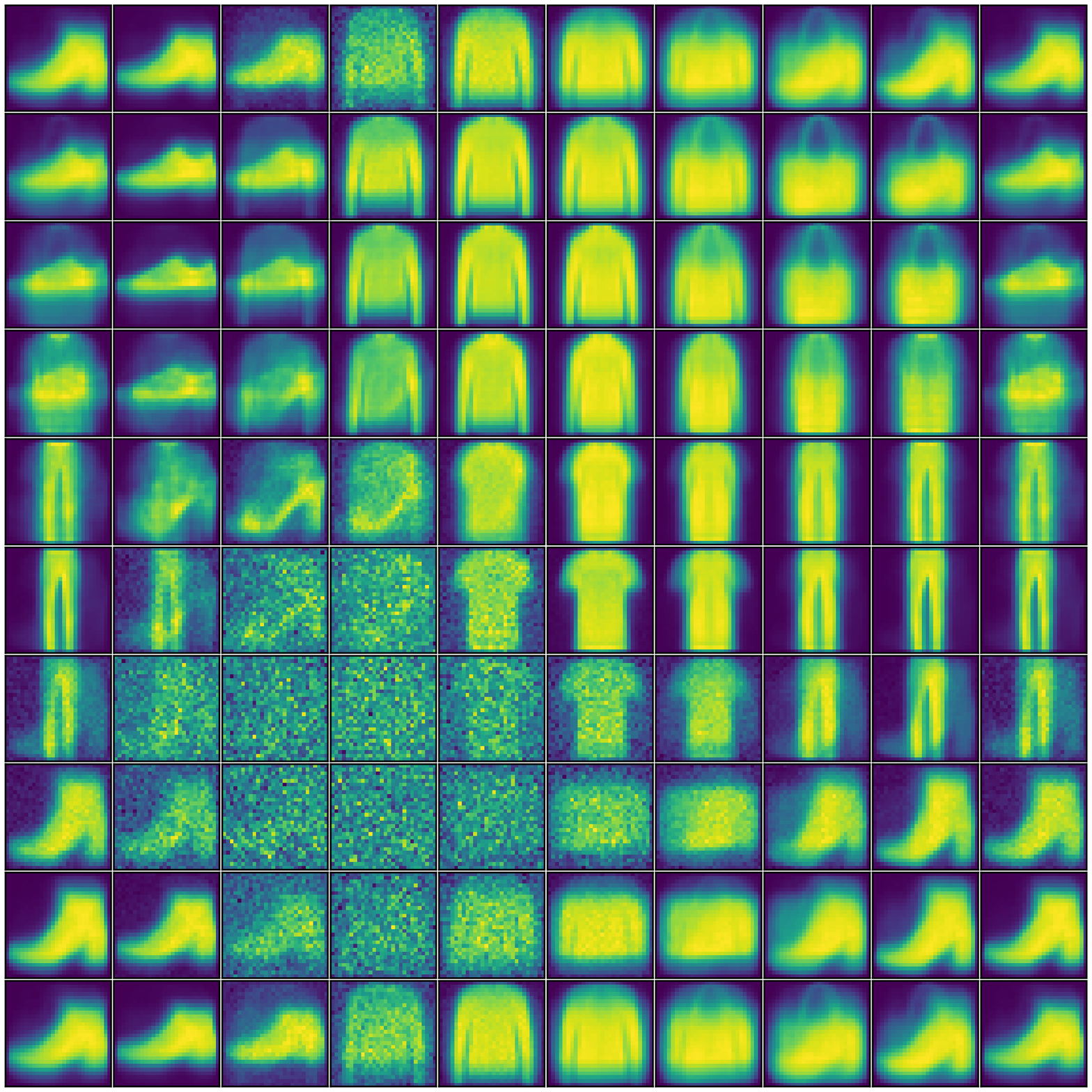}
  \caption{DSOM}
  \label{fig:w_dsom}
\end{subfigure}%
\begin{subfigure}{.5\textwidth}
  \centering
  \includegraphics[width=\linewidth]{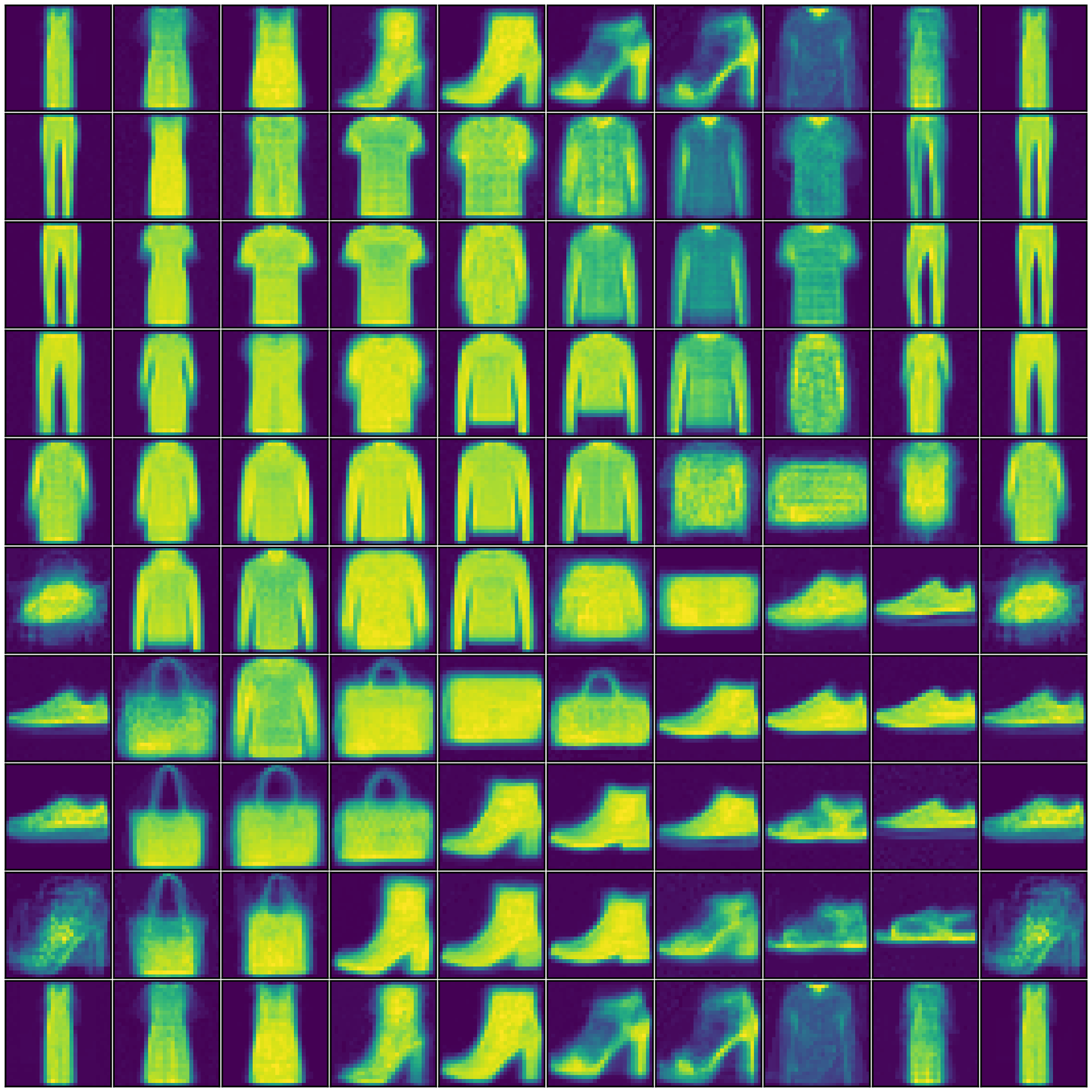}
  \caption{VDSOM}
  \label{fig:w_vbsom}
\end{subfigure}
\caption{Weights on the 2D grid}
\label{fig:w}
\end{figure}

The DSOM algorithm of \cite{rougier2011} is compared to ours (VDSOM) on $20000$
samples of the MNIST Fashion dataset on a $10 \times 10$ toroidal grid. DSOM
uses a learning rate of $\alpha=10^{-3}$ while VDSOM uses the same configuration
as before. In Fig. \ref{fig:sensitivity}, the distorsion is plotted as a
function of the elasticity $\eta$. VDSOM outperforms DSOM while being less
sensitive. In Fig. \ref{fig:w}, the weights of both methods trained with their
respective optimal elasticity are displayed on a grid. Note that some of the
DSOM weights are noise while this is not the case for VDSOM. VDSOM weights also
seem less blurred.

\section{Conclusion}
We proposed organisation conditions that yield a method for training SOM with
adaptative neighborhood radius in a variational Bayesian framework. This method
has been validated in a non-stationary setting and compared in an
high-dimensional setting with an other adaptative method.


\begin{footnotesize}


\bibliographystyle{unsrt}
\bibliography{ref.bib}

\begin{thebibliography}{1}

\bibitem{kohonen2012self}
T.~Kohonen.
\newblock {\em Self-Organizing Maps}.
\newblock Springer Series in Information Sciences. Springer Berlin Heidelberg,
  2012.

\bibitem{berglund2006}
Erik Berglund and Joaquin Sitte.
\newblock The parameterless self-organizing map algorithm.
\newblock {\em IEEE transactions on neural networks / a publication of the IEEE
  Neural Networks Council}, 17:305--16, 04 2006.

\bibitem{rougier2011}
Nicolas Rougier and Yann Boniface.
\newblock Dynamic self-organising map.
\newblock {\em Neurocomputing}, 74(11):1840--1847, 2011.

\bibitem{verbeek2005}
Jakob~J Verbeek, Nikos Vlassis, and Ben~JA Kr{\"o}se.
\newblock Self-organizing mixture models.
\newblock {\em Neurocomputing}, 63:99--123, 2005.

\bibitem{bishop1998}
Christopher~M Bishop, Markus Svens{\'e}n, and Christopher~KI Williams.
\newblock Gtm: The generative topographic mapping.
\newblock {\em Neural computation}, 10(1):215--234, 1998.

\bibitem{bishop2006pattern}
C.M. Bishop.
\newblock {\em Pattern Recognition and Machine Learning}.
\newblock Information Science and Statistics. Springer, 2006.

\bibitem{kingma2014}
Diederik~P Kingma and Jimmy Ba.
\newblock Adam: A method for stochastic optimization.
\newblock {\em arXiv preprint arXiv:1412.6980}, 2014.

\bibitem{scikit}
F.~Pedregosa, G.~Varoquaux, A.~Gramfort, V.~Michel, B.~Thirion, O.~Grisel,
  M.~Blondel, P.~Prettenhofer, R.~Weiss, V.~Dubourg, J.~Vanderplas, A.~Passos,
  D.~Cournapeau, M.~Brucher, M.~Perrot, and E.~Duchesnay.
\newblock Scikit-learn: Machine learning in {P}ython.
\newblock {\em Journal of Machine Learning Research}, 12:2825--2830, 2011.

\end{thebibliography}

\end{footnotesize}


\end{document}